\documentclass[runningheads]{llncs}

\usepackage{eccv}

\usepackage{eccvabbrv}
\usepackage{multirow}  
\usepackage{makecell}  
\usepackage{microtype} 
\usepackage{amsmath}
\usepackage{tablefootnote}
\usepackage[bottom]{footmisc}
\usepackage{graphicx}  
\usepackage{booktabs}
\usepackage{amsfonts,amssymb} 
\usepackage{orcidlink}
\usepackage{marvosym}
\usepackage{authblk}
\usepackage[accsupp]{axessibility}  
\usepackage{hyperref}
\hypersetup{
    colorlinks=true,
    linkcolor=blue,
    filecolor=blue,      
    urlcolor=blue,
    citecolor=cyan,
}

\makeatletter
\renewcommand{\maketag@@@}[1]{\hbox{\m@th\normalsize\normalfont#1}}
\makeatother
\usepackage{hyperref}

\newcommand{\yaom}[1]{{\color{black} #1}}

\begin{document}


\title{Integer-Valued Training and Spike-Driven Inference Spiking Neural Network for High-performance and Energy-efficient Object Detection}

\titlerunning{SpikeYOLO}

\author{Xinhao Luo\inst{1*}\orcidlink{0009-0008-3187-4612}
\and
Man Yao\inst{1}\thanks{Equal contribution.}\orcidlink{0000-0002-0904-8524} \and
Yuhong Chou\inst{2}\orcidlink{0009-0003-7788-7287} \and
Bo Xu\inst{1}\orcidlink{0000-0002-1111-1529} \and
Guoqi Li\inst{1}\orcidlink{0000-0002-8994-431X}\textsuperscript{(\Letter)}}

\authorrunning{Luo et al.}

\institute{
Institute of Automation, Chinese Academy of Sciences \linebreak
\email{\{luoxinhao2023, man.yao, guoqi.li\}@ia.ac.cn}\and
Xi'an Jiaotong University
}

\maketitle

\begin{abstract}
Brain-inspired Spiking Neural Networks (SNNs) have bio-plausibility and low-power advantages over Artificial Neural Networks (ANNs). Applications of SNNs are currently limited to simple classification tasks because of their poor performance. In this work, we focus on bridging the performance gap between ANNs and SNNs on object detection. Our design revolves around network architecture and spiking neuron. First, the overly complex module design causes spike degradation when the YOLO series is converted to the corresponding spiking version. We design a SpikeYOLO architecture to solve this problem by simplifying the vanilla YOLO and incorporating meta SNN blocks. Second, object detection is more sensitive to quantization errors in the conversion of membrane potentials into binary spikes by spiking neurons. To address this challenge, we design a new spiking neuron that activates Integer values during training while maintaining spike-driven by extending virtual timesteps during inference. The proposed method is validated on both static and neuromorphic object detection datasets. On the static COCO dataset, we obtain \textbf{66.2\%} mAP@50 and \textbf{48.9\%} mAP@50:95, which is \textbf{+15.0\%} and \textbf{+18.7\%} higher than the prior state-of-the-art SNN, respectively. On the neuromorphic Gen1 dataset, we achieve 67.2\% mAP@50, which is \textbf{+2.5\%} greater than the ANN with equivalent architecture, and the energy efficiency is improved by \textbf{5.7}$\times$. Code: \href{https://github.com/BICLab/SpikeYOLO}{https://github.com/BICLab/SpikeYOLO}

\end{abstract}
\renewcommand{\thefootnote}{}
\footnotetext[2]{(\Letter) Corresponding author.}
\renewcommand{\thefootnote}{\arabic{footnote}}


\section{Introduction}

\label{sec:intro}
Brain-inspired SNNs are known for their low power consumption\cite{roy2019towards,schuman2022opportunities}. Spiking neurons incorporate spatio-temporal information and emit spikes when the membrane potentials exceed a threshold\cite{maass1997networks}. Thus, spiking neurons trigger sparse additions only when they receive a spike and are otherwise idle. This spike-driven enables SNNs to exhibit obvious low-power advantages over ANNs when deployed on neuromorphic chips\cite{merolla2014million,davies2018loihi,pei2019towards,Speck}. However, the negative impact of complex neuronal dynamics and spike-driven nature is that SNNs are difficult to train and have limited task performance and application scenarios\cite{10242251}. 

For example, most applications of SNN algorithms in computer vision are limited to simple image classification tasks\cite{fang2021deep,Wang_2023_ICCV,10377657,guo2023rmp,yao2023spike,li2024seenn,hu2021advancing,deng2022temporal,hu2024highperformance,qiu2024gated}. Another more commonly used and challenging computer vision task, object detection, is rarely explored in SNNs. In 2020, Spiking-YOLO\cite{kim2020spiking} provided the first object detection model in deep SNNs, exploiting the method of converting ANN to SNN with thousands of timesteps. In 2023, EMS-YOLO\cite{su2023deep} became the first work to use direct training SNNs to handle object detection. Recently, the direct training Meta-SpikeFormer\cite{yao2024spikedriven} can process the object detection in a pre-training and fine-tuning manner for the first time. However, the performance gap between these works and ANNs is significant. In this work, we aim to bridge this gap and demonstrate the low-power of SNNs and their unique advantages in neuromorphic applications. We achieve this goal through two efforts.

First, we design a new architecture, SpikeYOLO, which combines the macro design of YOLO with the micro design of the SNN module. Simply replacing the artificial neurons in the YOLO series\cite{redmon2016you,bochkovskiy2020yolov4,wang2023yolov7} with spiking neurons generally does not work. Existing solutions include establishing the equivalence between ANN activation and spike firing rate\cite{kim2020spiking}, or improving the residual design\cite{su2023deep}. We argue that another potential reason is that the module design in the YOLO series is too complex, which is effective in ANNs, but not suitable for SNNs. We observe that after the complex YOLO modules are converted into the corresponding spiking versions, there is a phenomenon of spike degradation in which the deep layers almost no longer emit spikes. Therefore, we tend to simplify the design in SpikeYOLO. We only retain the macro architecture in YOLOv8\footnote{https://github.com/ultralytics/ultralytics} while using the simple meta SNN block in Meta-SpikeFormer\cite{yao2024spikedriven} as the basic module and performing micro design. 

Second, we design a novel spiking neuron, Integer Leaky Integrate-and-Fire (I-LIF), to reduce the quantization error of SNNs. Spike-driven is the key to low-power, while exploiting only binary signals will drop performance, especially in challenging object detection. Numerous studies have attempted to mitigate this problem, such as attention-based membrane optimization\cite{yao2023attention}, ternary spike\cite{guo2023ternary}, information max\cite{guo2022loss}, converting quantized ANNs\cite{hu2023fast}, optimal ANN2SNN\cite{bu2022optimal}. However, direct training based on optimization strategies can only alleviate errors; the optimal approximation of ANN2SNN requires large timesteps, and it is difficult to exploit the temporal information. In contrast, the idea of the proposed I-LIF is to use integer-valued activations to drop quantization errors and convert them into spikes during inference by extending the virtual timesteps. The features of I-LIF are: 1) Integer-valued training improves performance and is easy to train; 2) Integer activation during training can be equivalent to spike-driven in inference that there is only sparse addition; 3) The temporal dynamics of LIF are reserved, capable of processing neuromorphic object detection.

\yaom{The proposed method is validated on both static COCO\cite{lin2014microsoft} and neuromorphic Gen1\cite{de2020large} object detection datasets. The performance we report on both datasets far exceeds the existing best models in SNNs. In addition, we explore the performance changes after the mutual conversion of ANN and SNN with the same architecture. The main contributions of this work are:
\begin{itemize}
    \item \textbf{SpikeYOLO.} We explore suitable architectures in SNNs for handling object detection tasks and propose SpikeYOLO, which simplifies YOLOv8 and incorporates meta SNN blocks. This inspires us that the complex modules in ANN may not be suitable for SNN architecture design.
    
    \item \textbf{I-LIF Spiking Neuron.} We propose an I-LIF spiking neuron that combines integer-valued training with spike-driven inference. The former is used to reduce quantization errors in spiking neurons, and the latter is the basis of the low-power nature of SNNs.  
    
    \item \textbf{Performance.} The proposed method achieves outstanding accuracy with low power consumption on object detection datasets, demonstrating the potential of SNNs in complex vision tasks. On the COCO dataset, we obtain \textbf{66.2\%} mAP@50 and \textbf{48.9\%} mAP@50:95, which is \textbf{+15.0\%} and \textbf{+18.7\%} higher than the prior state-of-the-art SNN, respectively. On the Gen1 dataset, SpikeYOLO is \textbf{+2.5\%} better than ANN models with \textbf{5.7}$\times$ energy efficiency.
\end{itemize}
}

\section{Related Works}

\textbf{SNN Training Method.} Training methods have restricted the development of SNN for a long time. \yaom{To make SNNs deeper, two training methods have been developed.} ANN2SNN substitutes the ReLU function with \yaom{spiking} neurons, aiming to mimic the \yaom{continuous} activation by controlling the firing rate of \yaom{spiking} neurons\cite{cao2015spiking,diehl2015fast,sengupta2019going}. It can achieve high performance but often requires \yaom{long} timesteps and is difficult to process \yaom{sequence tasks that exploit the spatio-temporal dynamic nature of SNNs}. In contrast, another directly training an SNN leverage gradient \yaom{surrogate} to circumvent the non-differentiability of \yaom{binary spikes} \cite{wu2018spatio,neftci2019surrogate}. \yaom{Direct training is more flexible and} requires fewer timesteps, but its performance \yaom{usually suffers compared to ANNs of the same architecture.}  In this work, we focus on \yaom{utilizing directly trained SNNs to process object detection due to its more flexibility in architectural design.}

\textbf{SNN Architecture Design.} \yaom{The architecture of SNNs can be roughly divided into two categories: CNN-based and Transformer-based SNNs. Spiking ResNet has long dominated the SNN field because residual learning\cite{he2016deep} can address the performance degradation of SNNs as they become deeper. Typical spiking ResNet includes vanilla spiking ResNet\cite{zheng2021going}, SEW-ResNet\cite{fang2021deep}, and MS-ResNet\cite{hu2021advancing}. The main difference between them is the location of shortcuts and the ability to achieve identity mapping\cite{He_2016_identitymapping}. Recently, the Transformer\cite{transformer,dosovitskiy2021an} architecture has become popular \cite{mueller2021spiking,han2023complex,zhang2022spiking,zhang2022spike,zhang2022spike,zhou2023spikformer,yao2023spike,yao2024spikedriven} in the SNN field and has refreshed the performance upper bound. The SpikeYOLO we designed draws on the idea of the meta SNN block in Meta-SpikeFormer\cite{yao2024spikedriven} and merges it with the YOLOv8 architecture.} 

\textbf{Quantization Errors in SNNs.} Spiking neurons quantize continuous membrane potentials with complex spatio-temporal dynamics into binary spikes. \yaom{Obviously,} quantization error will limit the performance of SNNs. \yaom{In ANN2SNN, the reduction of quantization error is relatively simple, which can be achieved by increasing the timestep\cite{bu2022optimal}.} \yaom{In direct training,} adjusting the relationship between membrane potential distribution and threshold is the main solution, such as attention mechanism\cite{yao2023attention} optimizes membrane potential distribution \yaom{to reduce} noise spikes; information max\cite{guo2022loss} \yaom{is derived from the information entropy} angle to optimize spike firing. \yaom{However, optimization cannot change the inherently quantized error nature of binary spikes.} The solution of this work is to use integer values to reduce quantization errors during training and convert them to binary spikes to ensure spike-driven inference.

\textbf{Object Detection.} Object detection is a challenging yet pivotal task. \yaom{Existing object detection} frameworks can be simply categorized into: two-stage frameworks (RCNN series\cite{girshick2014rich,girshick2015fast,ren2015faster})  and single-stage frameworks (YOLO\cite{redmon2016you,bochkovskiy2020yolov4,wang2023yolov7}, Detr\cite{carion2020end,zhu2021deformable} series). The former generate region proposals before determining precise object locations and classifications, whereas the latter directly ascertain these elements, offering a swifter, more streamlined solution. \yaom{The object detection task has always been difficult for SNNs, and there are few related works.} Current works include ANN2SNN-based YOLO\cite{kim2020spiking,yuan2024trainable} and directly-trained spiking YOLO\cite{su2023deep,yao2024spikedriven}. \yaom{Their performance is poor, and it isn't easy to meet the needs of real scenarios.}

\section{Methods}
We exploit SpikeYOLO to process both static and neuromorphic object detection datasets. We first introduce how network inputs are unified. Then, the details of SpikeYOLO architecture and I-LIF spiking neuron are presented, respectively.

\subsection{Network Input}
The input of SNNs can be denoted as $X \in  \mathbb{R} ^{T\times C\times H \times W}$, where $T$ is the timestep, $C$ is the channel, $H \times W$ denote the spatial resolution.

\textbf{Static Image.} To leverage the spatio-temporal capabilities of SNNs, it is common practice that static images are repeated and utilized as input for each timestep $T$. This is called direct input encoding\cite{wu2019direct,kim2022rate}, where the first layer of spiking neurons in the network encodes the continuous values of the input into spike signals.

\textbf{Neuromorphic Event Stream.} Neuromorphic data (also known as event-based vision) are generated by a Dynamic Vision Sensor (DVS), which only generates spikes when the logarithmic change in light intensity at a pixel surpasses a predefined threshold. An event-based stream is characterized as $(x_n, y_n, t_n, p_n)$, with each event capturing spatial coordinates$ \left( x,y\right)$, timestamp $t$ and polarity $p$, where $p \in \{ -1,1 \}$ indicates whether light intensity has increased or decreased. Neuromorphic vision offers several advantages\cite{gallego2020event,Speck}, such as low resource requirements, high temporal resolution, and high robustness. The spike-driven nature of SNN makes it naturally suitable for processing event streams. The general strategy for neuromorphic pre-processing is to aggregate the event stream within a fixed time window into a frame format\cite{yao2023sparser,yao2021temporal,yao2023inherent}. In this work, we follow this operation. Specifically, the total input window length is $T \times dt$, where $dt$ and $T$ are temporal resolution and timestep, respectively. 

\begin{figure}[t]  
    \centering
    \includegraphics[width=0.85\linewidth]{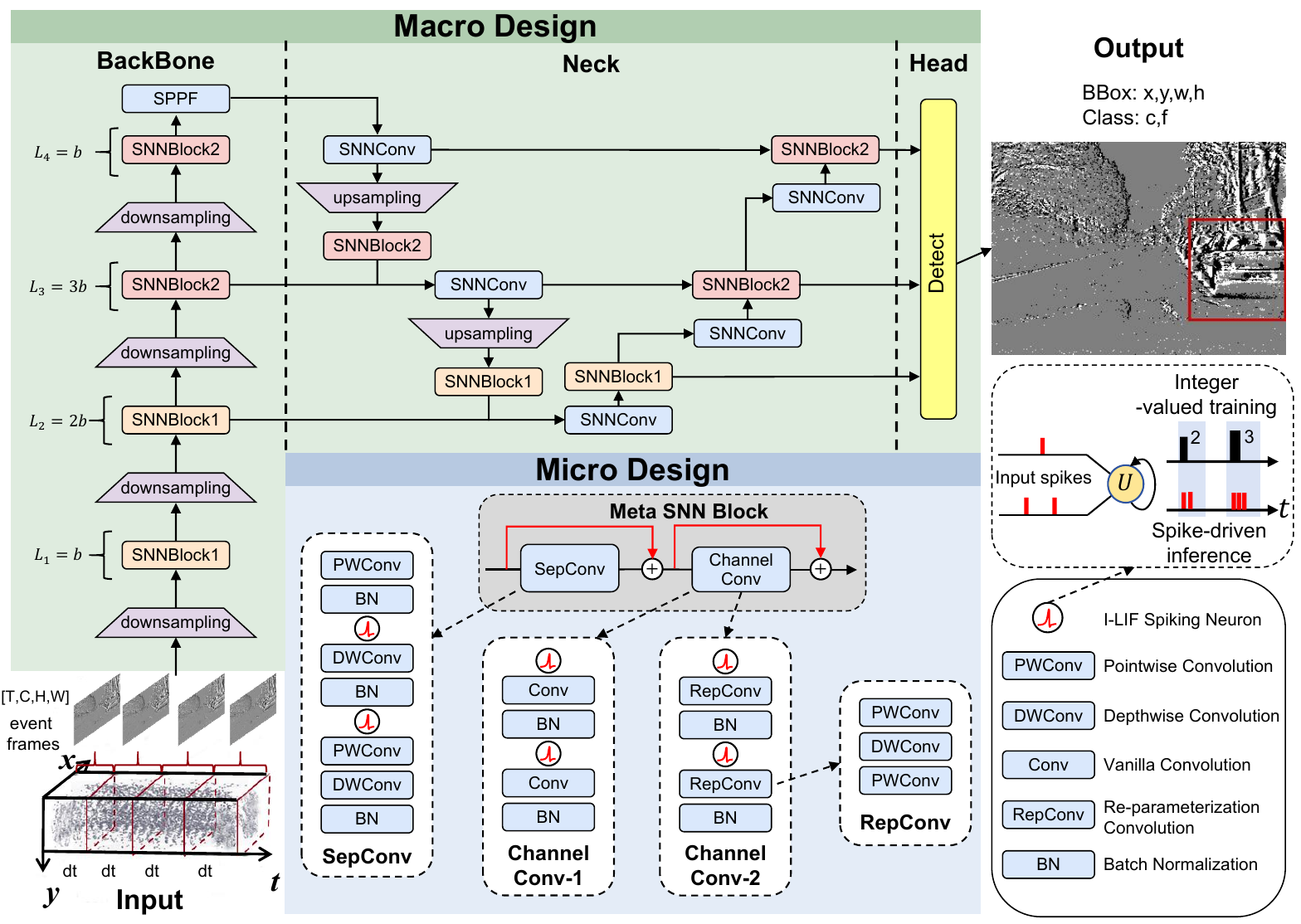} 
    \caption{The overall architecture of SpikeYOLO. We designed two SNN blocks, SNN-Block-1 and SNN-Block-2, and kept other architectures remain as YOLOv8. SNN-Block-1 employs standard convolution within its $\rm{ ChannelConv\left(\cdot\right)}$ component, whereas SNN-Block-2 utilizes re-parameterization convolution. That is, the difference between the two is the channel mixer module. In the low and high stages, we use SNN-Block-1 and SNN-Block-2, respectively. The spiking neuron is I-LIF, which activates integer values during training while converting them to binary spikes during inference.}
    \label{figure1}
    \end{figure}

\subsection{SpikeYOLO Architecture}
\textbf{Overview.} SpikeYOLO integrates the macro design of the YOLOv8 with the micro design of Meta-SpikeFormer\cite{yao2024spikedriven}. The motivation is that we observe complex computations within YOLO's modules result in spike degradation\cite{hu2021advancing} upon direct conversion to the SNN version. Consequently, we maintain the overarching design principles of the YOLO architecture while incorporating the inverted residual structure\cite{sandler2018mobilenetv2} and re-parameterization convolution\cite{ding2021repvgg} design of the Meta-SpikeFormer for detailed aspects.

\subsubsection{Network Output.} In object detection, the network outputs the class and position of each object based on the input image sequence $X = \{ X_t\}{^T_{t=1}} $. Suppose the input image sequence has $N$ goals, the output $B = \{ B_n\}{^N_{n=1}} $ can be calculated:

\begin{equation}  
 B = {\rm Model}\left(X\right),
\end{equation}
where each $B_n = \{ f_n,c_n,x_n,y_n,w_n,h_n\} $  contains information about degree of confidence $ f_n $, class $ C_n $, center coordinates $ \left(x_n,y_n\right) $ and target size $ \left(w_n,h_n\right) $. $ {\rm Model}\left(\cdot\right) $ refers to the proposed SpikeYOLO architecture.

\subsubsection{Macro Design.} Fig.~\ref{figure1} shows the overview of SpikeYOLO, a variation of the YOLO framework that is more suitable for the feature extraction scheme of SNNs. Specifically, YOLOv8 is a classic single-stage detection framework that partitions the image into numerous grids, with each grid responsible for predicting a target independently. Some classic designs, such as the feature pyramid network\cite{lin2017feature} in YOLOv8, play a crucial role in facilitating efficient feature extraction and fusion. By contrast, its feature extraction module, such as C2F, performs repeated feature extraction from the same set of feature maps. This module can enhance feature extraction in ANNs but does not work well in SNNs. As a compromise, we preserve the classic Backbone/Neck/Head architecture in YOLOv8 while incorporating strategies from the meta SNN block in Meta-SpikeFormer.

\subsubsection{Micro Design.} Meta-SpikeFormer\cite{yao2024spikedriven} is the current state-of-the-art architecture in SNNs, which explores the meta design of SNN and consists of CNN-based and Transformer-based SNN blocks. The meta block comprises a token mixer module and a channel mixer module. The difference between CNN-based and Transformer-based SNN blocks lies in the token mixer, which are spike-driven convolution and spike-driven self-attention, respectively. In this work, we mainly redesign the channel mixer module for the object detection task. As shown in Fig.~\ref{figure1}, SNN-Block-1 and SNN-Block-2 are designed to extract low-stage and high-stage features, respectively. 

The meta SNN block in \cite{yao2024spikedriven} can be written as:
\begin{equation}  
U' = U + {\rm SepConv} \left(U\right),
\end{equation}

\begin{equation}  
U'' = U' + {\rm ChannelConv} \left(U'\right),
\end{equation}
where $U \in  \mathbb{R}  ^{T\times C\times H \times W}$ is the layer input, ${\rm SepConv}\left( \cdot\right)$ is an inverted separable convolution module\cite{sandler2018mobilenetv2} with $ 7\times7 $ kernel size in MobileNetv2 to capture global features, followed by a $ 3\times3 $ depthwise convolution for further spatial feature fusion. ${\rm Sepconv}\left( \cdot\right)$ can be expressed as:

\begin{small} 
\begin{equation}  
{\rm SepConv} \left(U\right) = {\rm Conv_{dw2}}  \left( {\rm Conv_{pw2}}\left( SN \left({\rm Conv_{dw1}} \left(SN\left( {\rm Conv_{pw1}}\left( SN \left(U\right) \right) \right) \right)\right) \right) \right), 
\end{equation}

\end{small} 
where ${\rm Conv_{dw1}}\left(\cdot\right)$ and ${\rm Conv_{dw1}}\left(\cdot\right)$ are depthwise convolutions, ${\rm Conv_{pw1}}\left(\cdot\right)$ and ${\rm Conv_{pw1}}\left(\cdot\right)$ are pointwise convolutions\cite{8099678}. $SN (\cdot)$ is the spiking neuron layer. ${\rm ChannelConv}\left( \cdot\right)$ is the channel mixer, which facilitates inter-channel information fusion. We redesign ${\rm ChannelConv}\left(\cdot\right)$ in object detection task.

For SNN-Block-1, it is written as:
\begin{equation}  
{\rm ChannelConv1} \left(U'\right) =  {\rm Conv}  \left(SN\left( {\rm Conv}\left( SN \left(U'\right) \right) \right) \right),
\end{equation}

where ${\rm Conv\left( \cdot\right)}$ is a standard convolution with expansion ratio $r = 4$. In contrast, addressing high-stage features, SNN-Block-2 employs a re-parameterization convolution to minimize parameter count, which can be described as:

\begin{equation}  
{\rm ChannelConv2} \left(U'\right) =  {\rm RepConv}  \left(SN\left( {\rm RepConv}\left( SN \left(U'\right) \right) \right) \right), 
\end{equation}

\begin{equation}  
{\rm RepConv}  \left(U'\right) =  {\rm Conv_{pw2}}\left( {\rm Conv_{dw1}}\left( {\rm Conv_{pw1}} \left(U'\right) \right) \right), 
\end{equation}

where ${\rm RepConv}  \left(\cdot \right)$ is the re-parameterization convolution \cite{ding2021repvgg}  with kernel size $3 \times 3$, it can be re-parameterizated to a standard convolution during inference. 

\subsection{I-LIF Spiking Neuron}\label{I-LIF Spiking Neuron}
Spiking neurons propagate information in both spatial and temporal domains, and they mimic the spiking communication scheme of biological neurons. However, there are inherent quantization errors in converting the membrane potential of spiking neurons into binary spikes, which severely limits the representation of the model. Recently, Fast-SNN\cite{hu2023fast} achieves high-performance conversion with small timesteps by converting quantized ANNs into SNNs. This inspires us to ``why not train directly with integer values", which can significantly reduce the quantization error. We just need to be careful to ensure that the inference is spike-driven. So, we came up with the idea of I-LIF.

\subsubsection{LIF.} 
Leaky Integrate-and-Fire (LIF) spiking neuron\cite{maass1997networks} is the most popular neuron to construct SNNs due to its balance between bio-plausibility and computing complexity. The dynamics of LIF with soft reset is:

\begin{equation}  
 U\left[ t\right] = H\left[ t-1  \right] + X\left[ t \right], 
\end{equation}

\begin{equation}
\label{spike}
S\left[t\right] = \varTheta\left(U\left[t\right] - V_{th}\right),
\end{equation}

\begin{equation}
H\left[t\right] = \beta\left(U\left[t\right] - S\left[t\right]\right),
\end{equation}
where $t$ denotes the timestep, $U\left[ t\right]$ is the membrane potential that integrates the temporal information $H\left[ t-1\right]$ and spatial information $X\left[ t\right]$. $\varTheta\left(\cdot\right)$ is the Heaviside step function which equals 1 for $ x \geq 0$ and 0 otherwise. If $U\left[ t\right]$ exceeds the firing threshold $V_{th}$, spiking neuron fire a spike $S\left[ t\right]$ and $U\left[ t\right]$ will subtract it subsequently. Otherwise, $H\left[ t\right]$ will remain unchanged. And, $U\left[ t\right]$ decays to $H\left[ t\right]$ by a factor of ${\beta}$, which denotes the decay constant. For simplicity, we focus on Eq.\ref{spike} and denote the spiking neuron layer as $SN (\cdot)$, with its input as membrane potential tensor $U$ and its output as spike tensor $S$.

\subsubsection{I-LIF.}
\begin{figure}[t]  
    \centering
    \includegraphics[width=0.85\linewidth]{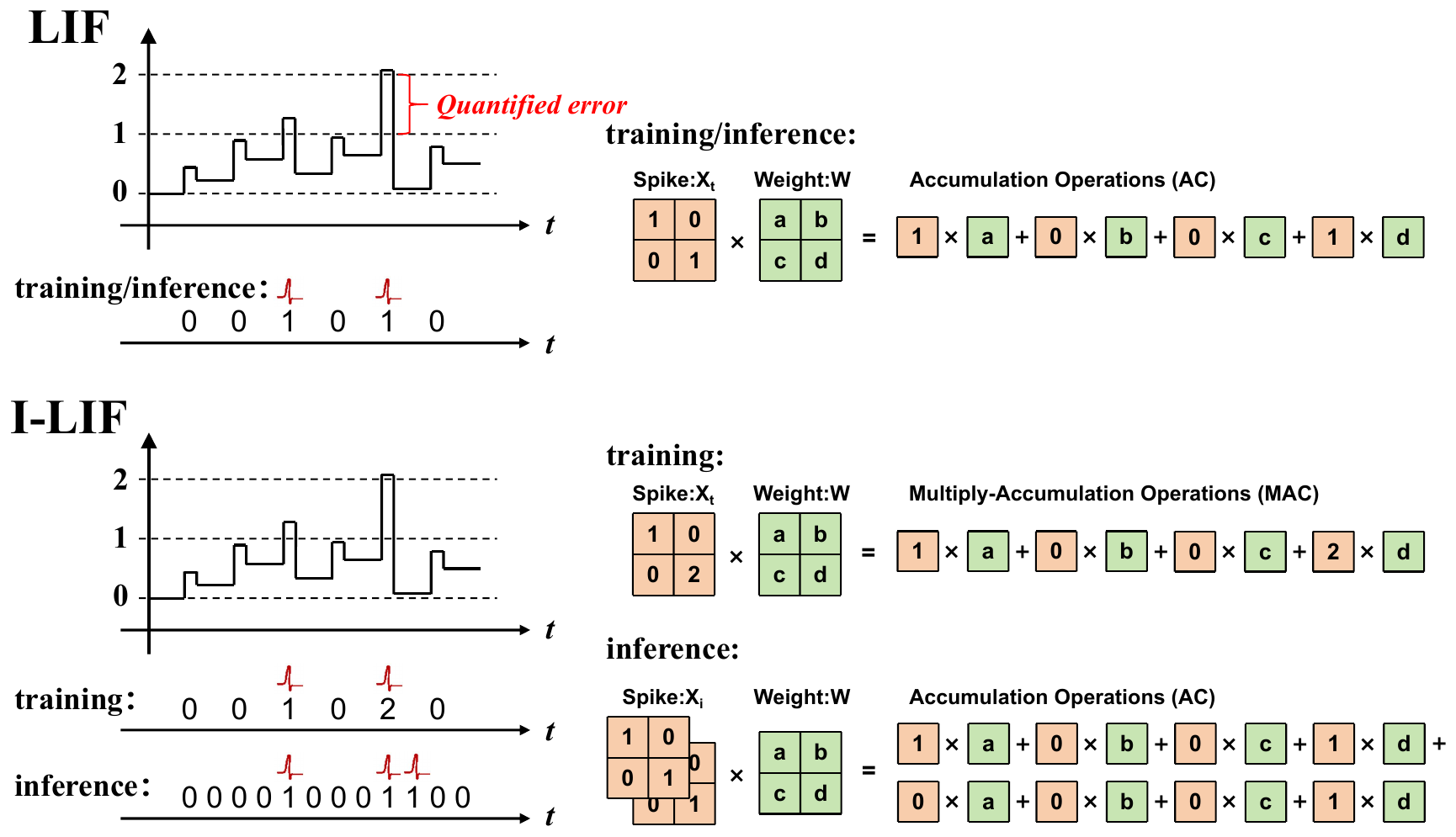} 
    \caption{Comparison of I-LIF and LIF. Binary spikes are emitted by LIF during both training and inference processes, which results in quantization errors. I-LIF emits integer values during the training process to reduce quantization errors, and converts them into binary spikes during inference to make the network only perform sparse addition.}
    \label{figure2} 
    \end{figure}

We propose the Integer Leaky Integrate-and-Fire (I-LIF) neuron to reduce the quantization error. As shown in Fig.~\ref{figure2}, I-LIF emits integer values while training, and converts them into 0/1 spikes when inference. Specifically, in I-LIF, Eq.\ref{spike} is rewritten as:
\begin{equation}
S\left[t\right] = Clip\left(round(U[t]), 0, D\right) \label{clip},
\end{equation}
where $round(\cdot)$ is a round symbol, $Clip\left(x,min,max\right)$ denotes that clipping $x$ to $[min, max]$, $D$ is a hyperparameter indicating the maximum emitted integer value by I-LIF.

\textbf{Training Stage.} Eq.\ref{clip} is not a continuous function, making its derivative a step function, potentially causing training instability. Previous studies have introduced several surrogate gradient functions, which primarily address binary spike outputs. We consistently utilize rectangular windows as the surrogate function. For simplicity, We retain gradients solely for neurons activated in the $[0, D]$ range, nullifying all others. 

\begin{figure}[ht]  
    \centering
    \includegraphics[width=0.85\linewidth]{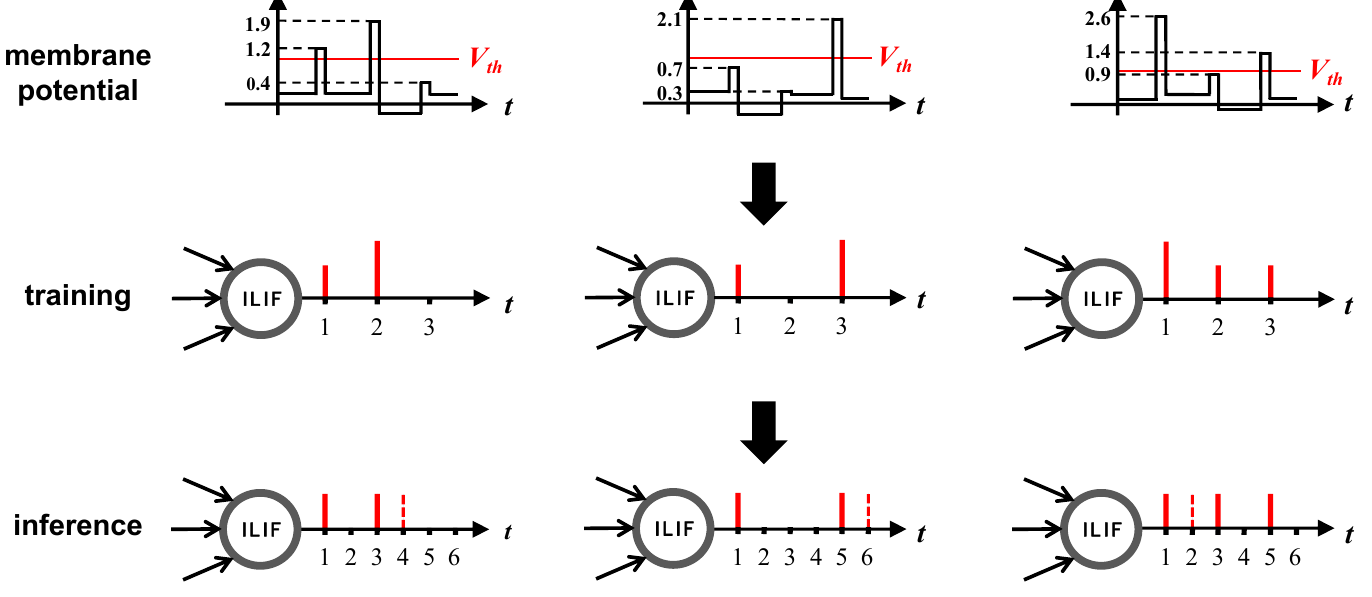} 
    \caption{An example of how the proposed I-LIF works. We assume $T=3$,$D=2$, and show the corresponding binary spike sequences of integer value during inference. The membrane potential in $\left[0.5,1.5\right)$ are quantized to 1, while those in $\left[1.5,2.5\right)$ are quantized to 2. membrane potential that $>2.5$ are also quantized to 2 due to the maximum integer value $D=2$. Subsequently, the membrane potential will be subtracted from the integer value. The training spike with a value of 2 will be converted into two binary spikes by extending virtual timesteps during inference.} 
    \label{figure3}
    \end{figure} 

\textbf{Inference Stage.} Introducing integer value necessitates additional MACs (Multiply-Accumulation operations), potentially diminishing the energy efficiency of SNNs. Thus, converting integer values to binary spikes is essential. Fig.~\ref{figure3} shows an example of how integer values convert to binary spikes by extending virtual timesteps during inference. Specifically, the input to the spiking neuron at $l+1$ layer can be described as $X^{l+1}\left[t\right] = W^lS^{l}\left[t\right]$. We extend the $T$ time step to $ T \times D$, and convert the integer value $S^l\left[t\right]$ to a spike sequence $\{ S^l\left[t,d\right]\}{^D_{d=1}} $, which satisfied:
\begin{equation}
 \sum_{d=1}^{D} S^l\left[t,d\right] =  S^l\left[t\right].
\end{equation}
Thus, the neuron's input at $l+1$ layer is reformulated as:
\begin{equation}
  X^l\left[t\right] = W^l\sum_{d=1}^{D} S^l\left[t,d\right].
\end{equation}
Given that matrix multiplication functions as linear operators, we establish:
\begin{equation}
W^l\sum_{d=1}^{D} S^l\left[t,d\right] =  \sum_{d=1}^{D} (W^lS^l\left[t,d\right]).
\end{equation}
Therefore, the input of the neuron at \textit{l}+1 layer can be computed by:
\begin{equation}
 X^l\left[t\right] =  \sum_{d=1}^{D} (W^lS^l\left[t,d\right]).
\end{equation}

The spike sequence $S^l\left[t,d\right]$ only contains 0/1, so all MACs can be converted into sparse ACs(Accumulation operations), which can ensure spike-driven when inference.

\section{Experiments}
We evaluate the proposed method on COCO 2017 val\cite{lin2014microsoft} and neuromorphic Gen1\cite{de2020large} datasets. The mean Average Precision(mAP) at IOU=0.5(mAP@50), the average mAP between 0.5 and 0.95(mAP@50:95), and energy cost are reported for each model. Specifically, the power of ANNs and SNNs can be calculated as:
\begin{equation}
\label{Eq:ANN_power}
E_{ANN} = O^2 \times C_{in} \times C_{out} \times k^2 \times  E_{MAC},
\end{equation}

\begin{equation} 
\label{Eq:SNN_power}
E_{SNN} = (T \times D) \times fr \times  O^2 \times C_{in} \times C_{out} \times k^2 \times E_{AC},
\end{equation}
where $O$ is the feature output size, $C_{in}$ and $C_{out}$ denotes the number of input channel and output channel, $k$ is the kernel size, $fr$ denotes the \yaom{average spike} firing rate, $T$ is the timestep, \yaom{$D$ is the upper limit of integer activation during training. We follow the most commonly used energy consumption evaluation method in the SNN field\cite{panda_2020_toward,yin2021accurate,yao2023attention}}. All operations assume a 32-bit floating-point implementation on 45nm technology, where $E_{MAC} = 4.6 pJ$ and $E_{AC} = 0.9 pJ$\cite{horowitz20141}. \yaom{As can be seen from Eq. \ref{Eq:ANN_power} and \ref{Eq:SNN_power}, SNN's low power comes from its sparse addition operation. The fewer spikes, the sparser the computation.}

\begin{table}[tb]
  \caption{Results on COCO 2017 val\cite{lin2014microsoft}. \yaom{$T \times D$ means that we set up $T$ timesteps, and each timestep is expanded $D$ times. In prior SNNs and ANNs, $D$ defaults to 1.}
  }
  \label{table1}
  \footnotesize
  \centering
  \tabcolsep=0.08cm  

  \begin{tabular}{@{}ccccccc@{}}  
    \toprule
    \multirow{2}{*}{Architecture} & \multirow{2}{*}{Model} & Param & Power & \multirow{2}{*}{$T \times D$}&  mAP@  & mAP@ \\  
                         &    & (M) & (mJ) &    &  50(\%)  & 50:95(\%) \\  
    
    \midrule   
    \multirow{3}{*}{ANN}       
    &PVT\cite{wang2021pyramid}   &  32.9 & 520.3 & 1 & 59.2 & 36.7 \\  
                               
    &DETR\cite{carion2020end}    &  41.0 & 197.8 & 1 & 62.4 & 42.0 \\  
    &YOLOv5 \tablefootnote{https://github.com/ultralytics/yolov5}  &  21.2 & 112.5 & 1 & 64.1 & 45.4 \\
    \Xcline{1-7}{0.06pt}  
    \multirow{3}{*}{ANN2SNN}& Spiking-Yolo\cite{kim2020spiking}  &  10.2 & -   & 3500 & - & 25.7\\ 
    &Bayesian Optim\cite{kim2020towards} &  10.2  & -& 5000  & -  &  25.9 \\  
    & Spike Calib\cite{li2022spike}  &  17.1 & -     & 512   &45.4   & -     \\

    \Xcline{1-7}{0.06pt}  
    \multirow{12}{*}{\makecell{Directly-trained\\ SNN}} & EMS-YOLO\cite{su2023deep}   
                        &  26.9 & 29.0  & 4     & 50.1  & 30.1 \\
    \Xcline{2-7}{0.06pt} 
                    & Meta-SpikeFormer                         
                        &  34.9 & 49.5  & 1     & 44.0  & -  \\ 
                    & (MaskRCNN)\cite{yao2024spikedriven}                         
                        &  75.0 & 140.8 & 1     & 51.2  & -  \\ 
    \Xcline{2-7}{0.06pt}   
                    & Meta-SpikeFormer  
                        &  16.8   & 34.8   & 1  & 45.0 & - \\
                    & (YOLO)\cite{yao2024spikedriven}
                        &  16.8   & 70.7 & 4 & 50.3 & - \\
                        
        \Xcline{2-7}{0.06pt}
                        & \multirow{7}{*}{\bf SpikeYOLO(Ours)}
                        &  13.2 & 23.1  & $1\times4$  & 59.2 & 42.5   \\ 
    \Xcline{3-7}{0.06pt} 
                    &   &  23.1 & 18.4  & $1\times1$  & 52.7 & 36.1  \\ 
                    &   &  23.1 & 34.6  & $1\times4$  & 62.3 & 45.5   \\ \Xcline{3-7}{0.06pt}
                    &   &  23.1 & 67.6  & $4\times1$  & 55.7 & 38.7   \\
                    &   &  23.1 & 134.7 & $4\times4$  & 63.3 & 46.3   \\ 
    \Xcline{3-7}{0.06pt} 
                    &   &  48.1 & 68.5  & $1\times4$  & 64.6 & 47.4  \\ 
    \Xcline{3-7}{0.06pt} 
                    &   &  68.8 & 84.2  & $1\times4$  & \textbf{66.2} & \textbf{48.9}  \\ 

  \bottomrule
  \end{tabular}  
\end{table}

\subsection{COCO 2017 val Dataset}   
\textbf{Experimental Setup.}  As a predominant static dataset for object detection, COCO 2017 val\cite{lin2014microsoft} comprises 80 classes split into 118K training and 5K validating images. In all experiments, we set decay factor $\beta = 0.25$, learning rate to 0.01, and adopt SGD optimizer. The models are trained for 300 epochs with a batch size of 40 on 4 NVIDIA V100 GPUs. Mosaic data augmentation\cite{bochkovskiy2020yolov4} technique is employed. The network structure is given in the supplementary material. Note, in our method, inference timestep is reported as $T \times D$, e.g., $1 \times 4$ denotes $T=1, D=4$.

\begin{table}[t]
  \caption{\yaom{Ablation studies of architectural design. We first convert YOLOv8 directly into the corresponding spiking version. Then we convert the SpikeYOLO designed in this work into the corresponding ANN version.}}
  \label{table2}
  \footnotesize
  \tabcolsep=0.08cm  
  \centering

  \begin{tabular}{@{}ccccccc@{}}  
    \toprule

    \multirow{2}{*}{Architecture} & \multirow{2}{*}{Model} & Param & Power & \multirow{2}{*}{$T \times D$}&  mAP@  & mAP@ \\  
                         &    & (M) & (mJ) &    &  50(\%)  & 50:95(\%) \\  
    \midrule   

    \multirow{2}{*}{ANN $\rightarrow$ SNN}   
        &YOLOv8  & 25.8 & 183.5 & 1 & 67.2 & 50.2 \\
        &Spiking YOLOv8    &  25.8 & 7.2 & $1 \times 1$ & 46.8 & 31.3 \\ 
    \Xcline{1-7}{0.06pt} 
    \multirow{2}{*}{SNN $\rightarrow$ ANN}   
        &SpikeYOLO (\textbf{Ours})  &  23.1 & 18.4 & $1 \times 1$  & 52.7  & 36.1 \\                         
        &YOLO  &  23.1 & 314.1  & 1 & 65.0 & 48.1  \\

  \bottomrule
  \end{tabular}  
\end{table}

\yaom{\textbf{Main Results} are shown in Table \ref{table1}. The proposed SpikeYOLO significantly improves the performance upper bound of the COCO dataset in SNNs. We obtain \textbf{66.2\%} mAP@50 and \textbf{48.9\%} mAP@50:95, which is \textbf{+15.0\%} and \textbf{+18.7\%} higher than the prior state-of-the-art SNN\cite{yao2024spikedriven}, respectively. SpikeYOLO also has significant advantages over existing SNNs in terms of parameters and power:} {\bf SpikeYOLO} vs. EMS-YOLO\cite{su2023deep}: Param, {\bf23.1M} vs. 26.9M; mAP@50, {\bf62.3}\% vs. 50.1\%; mAP@50:95, {\bf45.5}\% vs. 30.1\%; Power, {\bf33.2mJ} vs. 29.0mJ. \yaom{Moreover, the performance gap between SNNs and ANNs is significantly narrowed. For example, under similar parameters, the performance of SpikeYOLO and YOLO v5 are comparable, and the energy efficiency is $3.3\times$.}

\yaom{\textbf{Ablation Studies of Architectural Design.} We simplify YOLOv8 for SNN and incorporate meta SNN blocks. As shown in Table \ref{table1}, this architectural improvement enables the accuracy of spikeYOLO at $T=1$ and $D=1$ to reach 52.7\%, better than the prior state-of-the-art SNN. We are also interested in the question ``\emph{whether the architectures in SNNs and ANNs can be used directly interchangeably?}". We conduct the experiments in Table~\ref{table2}. We observe that directly converting the ANN architecture into the corresponding SNN brings significant performance degradation. The special architectural design of SNNs can improve its representation.}  

\yaom{\textbf{Ablation Studies of Quantization Error.} Integer-valued training is designed to reduce quantization error in SNNs. The larger $D$ is, the smaller the quantization error is. In Table \ref{table1}, we fixed the parameters to 23.1M. When $T=1$ and $T=4$, we expand $D=1$ to $D=4$, respectively, and the accuracies of mAP@50 are increased by +9.6\% and +7.6\%. In contrast, if we fix $D=1$ and increase $T=1$ to $T=4$, the performance improvement of mAP@50 is only 3\%. These results show that quantization error has a greater impact on performance than the setting of timesteps. And, in terms of power, increasing $D$ is more cost-effective than increasing $T$. For instance, when $1 \times 1$ changes to $1 \times 4$, power increases by 88\%; while $1 \times 1$ changes to $4 \times 1$, power improves by 267\%.}

\begin{figure}[t]  
    \centering
    \includegraphics[width=0.8\linewidth]{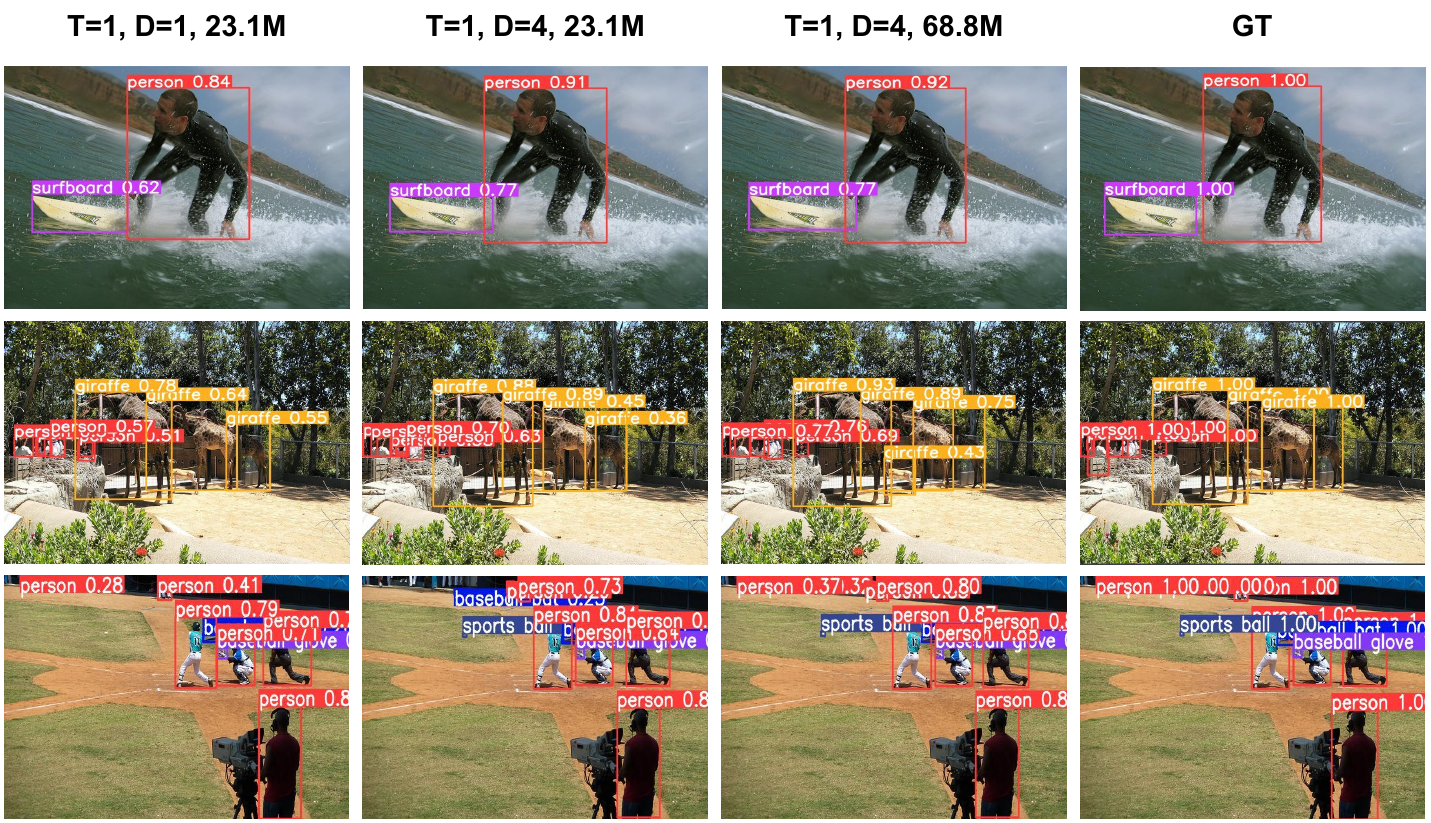} 
    \caption{The object detection results on the COCO dataset. The first two columns compare the effect of maximum integer value $D$ on performance for the same structure. The second and third columns compare the effect of the size of the model on performance.} 
\label{figure4}
\end{figure}

\subsection{Gen1 Automotive Detection Dataset}

\yaom{\textbf{Experimental Setup.}} As a large neuromorphic object detection dataset, Gen1\cite{de2020large} encompasses 39 hours of open road and various driving scenarios, captured using an ATIS sensor with a resolution of $304\times240$ pixels. The dataset is organized into training, validation, and testing subsets. The bounding box annotations of pedestrians and cars(over 255,000) were manually labeled. For each annotation, we process the event-based stream 2.5 seconds before its occurrence, dividing it into $T$ slices for model input. We train the model for 50 epochs and maintain other hyperparameters same as the COCO 2017 dataset. 

{\textbf{Main Results} \yaom{on Gen1 dataset} are shown in Table~\ref{table3}. The proposed SpikeYOLO notably elevates the performance benchmark for the Gen1 dataset in SNNs. We achieve \textbf{67.2\%} mAP@50 with 23.1M parameters, which outperforms the prior state-of-the-art SNN model by \textbf{+8.2\%}. For example, when $T = 5$, {\bf SpikeYOLO} vs. EMS-YOLO\cite{su2023deep}: Param, {\bf13.2M} vs. 14.4M; mAP@50, {\bf66.0}\% vs. 59.0\%; mAP@50:95, {\bf38.5}\% vs. 31.0\%. In contrast to the COCO dataset, Gen1 contains temporal information that is more suitable for SNN processing. We conduct experiments on the performance of SNN and ANN with the same architecture. We observe that SpikeYOLO's mAP@50 accuracy is \textbf{+2.5\%} higher than the corresponding ANN, and shows a {\bf5.7}$\times$ energy efficiency. This indicates that SNN has attractive potential in processing neuromorphic data.

\textbf{Ablation Studies of Quantization Error.} Both $T$ and $D$ significantly influence outcomes when processing neuromorphic datasets. Table~\ref{table4} gives a comprehensive ablation study on SpikeYOLO with 23.1M parameters that evaluate the effects of varying $T$ and $D$. We observe some interesting experimental results. \emph{First}, boosting the timestep $T$ will bring improvement in accuracy and power. For instance, with the set of $D=1$, extending $T=1$ to $T=4$ yields a {\bf+6.7\%} increase in mAP@50, and the power will increase by $3.7\times$. But further extending $T=4$ to $T=8$ results in a marginal increase of only +0.7\% and significantly increases energy cost. \emph{Second}, We were surprised to see that by expanding $D$ at a fixed $T$, the performance will be improved, while the power will be dropped. For example, $2 \times 1$ vs. $2 \times 2$ vs. $2 \times 4$: mAP@50, 63.6\% vs. 66.1\% vs. 67.0\%; Power, 8.1mJ vs. 7.8mJ vs. 7.1mJ. \textbf{This trend is completely different from SpikeYOLO’s performance in COCO,} where the increase of $D$ will bring more energy cost. We argue that this phenomenon is because SNN exhibits various spike firing for dense/sparse data.

\begin{table}[t]
  \caption{Results on the Gen1 dataset\cite{de2020large}. * We convert 23.1M of SpikeYOLO into ANN with the same architecture. 
  }
  \label{table3}
  \footnotesize
  \centering
  \tabcolsep=0.08cm  
  \begin{tabular}{@{}ccccccc@{}}  
    \toprule

    \multirow{2}{*}{Architecture} & \multirow{2}{*}{Model} & Param & Power & \multirow{2}{*}{$T \times D$}&  mAP@  & mAP@ \\  
                         &    & (M) & (mJ) &    &  50(\%)  & 50:95(\%) \\  
    \midrule   
    \multirow{2}{*}{ANN} 
    
    &YOLOv3-tiny\cite{redmon2018yolov3}
    &  10.2 &  5.1  & 1  & 44.5 & -         \\   
    &SpikeYOLO*   
    &  23.1 & 73.5 & 1  & 64.7 & 39.7      \\
    \Xcline{1-7}{0.06pt}  

    \multirow{11}{*}{SNN}          
    & \multirow{3}{*}{EMS-YOLO\cite{su2023deep}}     
    &  6.2  & 1.2 & 5 & 54.7 & 26.7 \\
    &                           
    &  9.3  & 2.0 & 5 & 56.5 & 28.6 \\
    &                               
    &  14.4 & 3.4 & 5 & 59.0 & 31.0 \\
    \Xcline{2-7}{0.06pt} 
    & VGG-11+SDD                    
    &  12.6 &  11.1  & 1 & - & 17.4 \\ 
    & MobileNet-64+SSD              
    &  24.3 &  5.7  & 1 & - & 14.7\\ 
    & DenseNet121-24+SSD\cite{cordone2022object}
    & 8.2   &  3.9  & 1 & - & 18.9  \\ 
        \Xcline{2-7}{0.06pt}  
    & Spiking-Yolo\cite{kim2020spiking}
    & 7.9 &  102.3  & 500 & 44.2 & - \\  
    & Tr-Spiking-Yolo\cite{yuan2024trainable}
    & 7.9 &  0.9  & 5   & 45.3 & - \\ 
        \Xcline{2-7}{0.06pt}  
    &\multirow{3}{*}{\bf SpikeYOLO(Ours)}
    &  13.2 &  11.0  &  $5\times1$  & 66.0 & 38.5 \\ 
    &                                       
    &  23.1 &  19.7  &  $5\times1$  & 66.4 & 38.9  \\ 
    &                                      
    &  23.1 &  12.9  &  $4\times2$  & \textbf{67.2} & \textbf{40.4} \\ 

  \bottomrule
  \end{tabular}  
\end{table}

\begin{table}[t]
  \caption{The influence of $T$ and $D$ on Gen1. We set SpikeYOLO (23.1M) as the baseline and vary $T$ and $D$ for each study.}

  \label{table4}
  \footnotesize
  \centering
  \tabcolsep=0.08cm  
  \begin{tabular}{@{}ccccc@{}}  
  
    \toprule
    Method & $T \times D$   & Power(mJ) &  mAP@50(\%)  & mAP@50:95(\%)\\  
    \midrule   
    \multirow{8}{*}{SpikeYOLO} 
    & $1 \times 1$  & 4.0     & 59.3   & 33.1 \\  
    & $1 \times 4$  & 3.9 (-0.1)     & 65.1 (+5.8)   & 38.9 (+5.8)\\  
    \Xcline{2-5}{0.06pt}
    & $2 \times 1$  &  8.1    & 63.6   & 36.5 \\  
    & $2 \times 2$  &  7.8 (-0.3)   & 66.1 (+2.5)   & 39.0 (+2.5)\\ 
    & $2 \times 4$  &  7.1 (-1.0)   & 67.0 (+3.4)  & 40.1 (+3.6)\\ 
    \Xcline{2-5}{0.06pt}
    & $4 \times 1$  &  14.8   & 66.0   & 38.4 \\  
    & $4 \times 2$  &  12.9 (-1.9)  & 67.2 (+1.2)   & 40.4 (+2.0) \\
    \Xcline{2-5}{0.06pt}
    & $8 \times 1$  &  27.0   & 66.7   & 39.3 \\  

  \bottomrule
  \end{tabular}  
\end{table}

\subsection{Architecture Ablation Experiments}

\textbf{Re-parameterization Design.} As shown in Table~\ref{table5}, if we remove re-parameteri-zation by adding neurons into inverted separable convolutions, the mAP@50 and mAP@50:95 will decrease 1.7\% and 1.8\% respectively.
 
\textbf{SNN Block Design.} Including a $3\times3$ standard convolution within the initial stages of convolution blocks is crucial. As shown in Table~\ref{table5}, substituting SNN-Block-1 for SNN-Block-2 leads to a reduction in performance of around 1\%. Moreover, we try to replace high-stage SNN-Block-2 with meta Transformer-based SNN block, just like Meta-SpikeFormer\cite{yao2024spikedriven}. We find that there is little to no performance gain by doing this and that the parameters increase. Therefore, only spiking CNN blocks are exploited in our SpikeYOLO.

\textbf{Detection Head.} The detection mechanisms within YOLO are categorized into anchor-based heads(\textit{e.g.}, YOLOv5) and anchor-free heads(\textit{e.g.}, YOLOv8). The former directly predicts each bounding box's dimensions, whereas the latter estimates the probability distribution of each bounding box. Previous EMS-YOLO\cite{su2023deep} and Meta-SpikeFormer\cite{yao2024spikedriven} employ anchor-based heads. SpikeYOLO exploits the anchor-free head because of its higher accuracy (see Table~\ref{table5}).

\begin{table}[tb]
  \caption{Ablation studies of architecture design. We set $T\times D =1\times 4$ and modify just one point of baseline to test how the parameters, power and performance vary.
  }

  \label{table5}
  \footnotesize
  
  \centering
  \tabcolsep=0.011cm  
  \begin{tabular}{@{}cccccc@{}}  
  
   \toprule
    Method & Param(M) &  mAP@50(\%)  & mAP@50:95(\%)\\  
    \midrule   
    SpikeYOLO(Baseline)          & 23.1     &  62.3  & 45.5 \\  
    Remove re-parameterization           &  23.1   & 60.6 & 43.7 \\  
    SNN-Block-1 → SNN-Block-2              &  19.9   & 61.2 & 44.7 \\  
    SNN-Block-2 → Transformer Block        &  24.5   & 61.0 & 44.2 \\  
    Anchor-free head → Anchor-based head &  21.2   & 59.5 & 39.7 \\  
  \bottomrule
  \end{tabular}  
\end{table}

\section{Conclusion}
\yaom{This work significantly narrows the performance gap between SNNs and ANNs on object detection tasks. We achieve this through network architecture and spiking neuron design. The proposed SpikeYOLO architecture abandons the complex module design in the vanilla YOLO series and exploits simple meta spike blocks to build the model. Then, the I-LIF spiking neuron capable of integer-valued training and spike-driven inference is proposed to drop quantization errors. We improve the upper bound of the SNN domain's performance on the COCO dataset by +15.0\% (mAP@50) and +18.7\% (mAP@50:95), respectively. On the neuromorphic Gen1 dataset, SpikeYOLO achieves better performance and lower power than ANN of the same architecture. Furthermore, we investigate the performance of equivalent architecture ANNs and SNNs in different datasets, and the results show that the redesigned SNN architecture performed better. This work enables SNNs to handle complex object detection and can inspire the application of SNNs in more visual scenarios.}

\clearpage 

%
%

\section*{Acknowledgements}

This work was partially supported by National Distinguished Young Scholars (62325603), and National Natural Science Foundation of China (62236009, U22A20103,62441606), Beijing Natural Science Foundation for Distinguished Young Scholars (JQ21015), National Science Foundation for Young Scientists of China, China Postdoctoral Science Foundation (GZB20240824, 2024M753497), and CAAI-MindSpore Open Fund, developed on OpenI Community.

\bibliographystyle{splncs04}
\bibliography{main}

\clearpage

\end{document}